\documentclass[runningheads]{llncs}
\usepackage{graphicx}

\usepackage{tikz}
\usepackage{comment}
\usepackage{amsmath,amssymb} 
\usepackage{color}

\usepackage{multirow}
\usepackage{subfigure}
\usepackage{booktabs}
\usepackage{wrapfig}

\usepackage[colorlinks, linkcolor=red, anchorcolor=red, citecolor=green]{hyperref}

\newcommand{\etal}{ \emph{et al.} }
\newcommand{\eg}{ \emph{e.g.}, }
\newcommand{\ie}{ \emph{i.e.}, }

\makeatletter
\newcommand{\rmnum}[1]{\romannumeral #1}
\newcommand{\Rmnum}[1]{\expandafter\@slowromancap\romannumeral #1@}
\makeatother

\begin{document}
\pagestyle{headings}
\mainmatter
\def\ECCVSubNumber{123}  

\title{Adversarial Self-Supervised Learning for Semi-Supervised 3D Action Recognition} 

\titlerunning{Adversarial Self-Supervised Learning}
%

\author{Chenyang Si\inst{1,2,3}\orcidID{0000-0002-3354-1968} \and
Xuecheng Nie\inst{3} \and
Wei Wang\inst{1,2} \and
Liang Wang\inst{1,2} \and
Tieniu Tan\inst{1,2} \and
Jiashi Feng\inst{3} 
}
\authorrunning{Chenyang Si et al.}
%
\institute{University of Chinese Academy of Sciences \and
CRIPAC \& NLPR, Institute of Automation, Chinese Academy of Sciences \and
Department of ECE, National University of Singapore\\
\email{chenyang.si@cripac.ia.ac.cn, \{wangwei, wangliang, tnt\}@nlpr.ia.ac.cn, niexuecheng@u.nus.edu, elefjia@nus.edu.sg}
}

\maketitle

\begin{abstract}

We consider the problem of semi-supervised 3D action recognition which has been rarely explored before. Its major challenge lies in how to effectively learn motion representations from unlabeled data. Self-supervised learning (SSL) has been proved very effective at learning representations from unlabeled data in the image domain. However, few effective self-supervised approaches exist for 3D action recognition, and directly applying SSL for semi-supervised learning suffers from misalignment of representations learned from SSL and supervised learning tasks. To address these issues, we present Adversarial Self-Supervised Learning (ASSL), a novel framework that tightly couples SSL and the semi-supervised scheme via neighbor relation exploration and adversarial learning. Specifically, we design an effective SSL scheme to improve the discrimination capability of learned representations for 3D action recognition, through exploring the data relations within a neighborhood. We further propose an adversarial regularization to align the feature distributions of labeled and unlabeled samples. To demonstrate effectiveness of the proposed ASSL in semi-supervised 3D action recognition, we conduct extensive experiments on NTU and N-UCLA datasets. The results confirm its advantageous performance over state-of-the-art semi-supervised methods in the few label regime for 3D action recognition.

\end{abstract}

\section{Introduction}
\label{Introduction}

Recently, 3D action recognition (a.k.a. skeleton-based action recognition) has made remarkable progress through learning discriminative features with effective networks \cite{Du2015Hierarchical,Zhang2017View,Chao2018Co-occurrence,Ke2017A,Yan2018Spatial,Shi_2019_CVPR,Si_2019_CVPR}. 
However, these methods heavily rely on the available manual annotations that are costly to acquire. 
Techniques requiring less or no manual annotations are therefore developed, and among them a powerful approach is semi-supervised learning. 
It is aimed at leveraging unlabeled data to enhance the model's capability of learning and generalization such that the requirement for labeled data can be alleviated. 
It has been widely applied in the image domain \cite{Durk2014Semi-supervised,Augustus2016Semi-supervised,Tim2016Improved,Samuli2017Temporal,Antti2017Mean,Takeru2018Virtual,Dong2013Pseudo}. 
Compared with these methods, \cite{Xiaohua2019S4l} has recently proposed a more efficient way of feature learning from unlabeled data, namely self-supervised semi-supervised learning ($S^4$L), that couples self-supervision with a semi-supervised learning algorithm. It employs the self-supervised technique to learn representations of unlabeled data to benefit semi-supervised learning tasks. Self-supervised learning is very advantageous in making full use of unlabeled data, which learns the representations of unlabeled data via defining and solving various pretext tasks. Thus in this work we exploit its application to semi-supervised 3D action recognition, which has little previous investigation.

\begin{wrapfigure}{R}{0.5\textwidth}
  \centering
    \includegraphics[width=0.95\linewidth, height=0.5\linewidth]{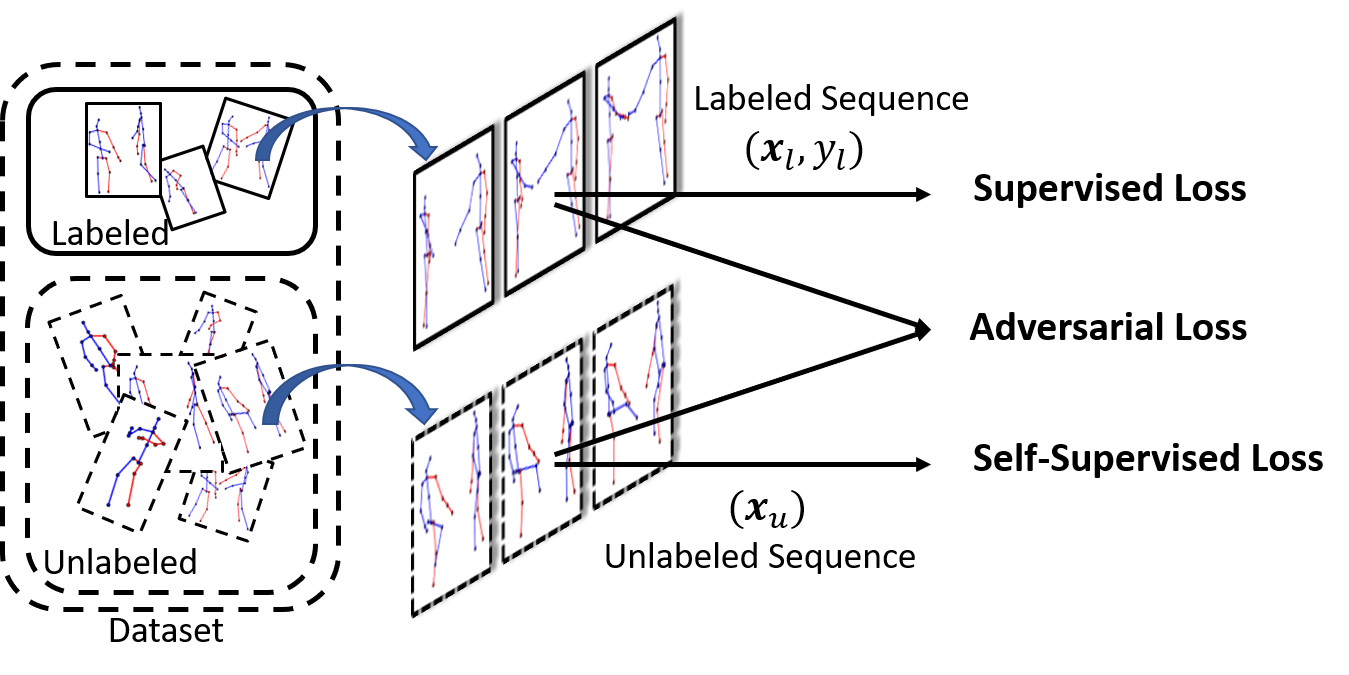}
 \caption{Illustration of our main idea. We design an effective SSL scheme to capture the discriminative motion representations of unlabeled skeleton sequences for 3D action recognition. Since directly applying SSL to semi-supervised learning suffers from misalignment of representations learned from SSL and supervised learning tasks, we further pioneer to align their feature distributions via adversarial learning
  }
  \label{fig_AdvsemiNet}
\end{wrapfigure}
As images contain rich information that is beneficial to feature extraction, many effective SSL techniques \cite{Alexey2014Discriminative,Raviteja2016Rolling,Zhirong2018Unsupervised} are image-based. Comparatively, for tasks over skeleton data which represent a person by 3D coordinate positions of key joints, it becomes very challenging to leverage SSL techniques to learn discriminative motion representation. Therefore, how to learn motion representation with SSL technique is an urgent problem for this task. Recently, \cite{Nenggan2018Unsupervised} proposes a SSL method to learn temporal information of unlabeled sequence via skeleton inpainting. This SSL treats each sample as an \emph{individual} such that it ignores the shared information among samples with the same action class. As a result, semi-supervised 3D action recognition has derived little benefit from the representations learned by skeleton inpainting.

Moreover, we also find that, directly applying SSL for semi-supervised learning suffers from misalignment of representations learned from self-supervised and supervised learning tasks. As shown in Fig.~\ref{fig_AdvsemiNet}, labeled and unlabeled samples are enforced with supervised and self-supervised optimization objectives respectively. Though both sampled from the same data distribution, their feature distributions are misaligned. This misalignment would weaken the generalization of semi-supervised 3D action recognition models to unseen samples.
A task with similar problem as ours is unsupervised domain adaptation (UDA) that matches the feature distributions from \emph{different domains}. While their problem is quite similar to ours, there exist important differences between UDA and our task. In UDA, the discrepancy of feature distributions is caused by different domains. Our problem is the misalignment of representations learned from SSL and supervised learning tasks in semi-supervised 3D action recognition. One line of research in UDA is adversarial-based adaptation methods \cite{Yaroslav2015Unsupervised,Eric2017Adversarial,Mingsheng2018Conditional} that have shown promising results in domain adaptation. These methods seek to minimize an approximate domain discrepancy distance through an adversarial objective with respect to a domain discriminator. Hence, inspired by the alignment effect of adversarial learning in UDA, we exploit its application to couple the self-supervision method into a semi-supervised learning algorithm.

In this work, we propose an Adversarial Self-Supervised Learning (ASSL) Network for semi-supervised 3D action recognition. 
As shown in Fig. \ref{fig_AdvsemiNet}, our model leverages (\rmnum{1}) self-supervised learning to capture discriminative motion representation of unlabeled skeleton sequences, and (\rmnum{2}) adversarial regularization that allows to align feature distributions of labeled and unlabeled sequences. More specifically, in addition to a self-inpainting constraint \cite{Nenggan2018Unsupervised} for learning temporal information of each \emph{individual} unlabeled sample, we propose a new perspective of consistency regularization within the neighborhood to explore the data relationships. Neighborhoods can be considered as tiny sample-anchored clusters with high compactness and class consistency. Consistency regularization within the neighborhood further reveals the underlying class concept of the self-supervised motion representation. Such discriminative motion representations significantly improve the performance of semi-supervised 3D action recognition. Moreover, considering that adversarial learning can minimize the discrepancy between two distributions, we also propose a novel adversarial learning strategy to couple the self-supervision method and a semi-supervised algorithm. The adversarial regularization allows the model to align the feature distributions of labeled and unlabeled data, which boosts the capability of generalization to unseen samples for semi-supervised 3D action recognition.

We perform extensive studies for semi-supervised 3D action recognition on two benchmark datasets: NTU RGB+D \cite{Shahroudy2016NTU} and N-UCLA \cite{Jiang2014Cross} datasets. With the proposed ASSL network, we establish new state-of-the-art performances of semi-supervised 3D action recognition. Summarily, our main contributions are in three folds:
\begin{enumerate}
    \item We present an Adversarial Self-Supervised Learning (ASSL) framework for semi-supervised 3D action recognition, which tightly couples SSL and a semi-supervised scheme via adversarial learning and neighbor relation exploration.  
    
   \item We offer a new self-supervised strategy, \ie neighborhood consistency, for semi-supervised 3D action recognition. By exploring data relationships within the neighborhood, our model can learn discriminative motion representations that significantly improve the performance of semi-supervised 3D action recognition. 
  
   \item We identify that directly applying SSL for semi-supervised learning suffers from the representation misalignment of labeled and unlabeled samples. A novel adversarial regularization is pioneered to couple SSL into a semi-supervised algorithm to align their feature distributions, which further boosts the capability of generalization.

\end{enumerate}

\section{Related Work}

\subsection{3D Action Recognition}

Human action recognition is one of important computer vision tasks. Due to the informative representation for the action, skeleton-based action recognition has been examined thoroughly in past literature. Previously, the traditional approaches \cite{Raviteja2014Human,Raviteja2016Rolling,Hussein2013Human,Wang2012Mining} try to design various hand-crafted features from skeleton sequences to represent human motion, \eg relative 3D geometry between all pairs of body parts~\cite{Raviteja2014Human}. 
Recently, deep learning has also been applied to this task due to its wide success. 
To model temporal dependencies, many methods leverage and extend the recurrent neural networks (RNNs) to capture the motion features for skeleton-based action recognition, \eg HBRNN \cite{Du2015Hierarchical}  and VA-LSTM \cite{Zhang2017View}. 
Based on Convolutional Neural Networks (CNNs) that are powerful at learning hierarchical representations, spatio-temporal representations are extracted for action recognition in \cite{Yong2015Skeleton,Ke2017A,Chao2018Co-occurrence,Pichao2016Action}. 
For graph-structured data, graph-based approaches \cite{Chenyang2018Skeleton,Li_2019_CVPR,Chenyang2020Skeleton-Based} are popularly adopted for skeleton-based action recognition, \eg ST-GCN \cite{Yan2018Spatial} and AGC-LSTM \cite{Si_2019_CVPR}. 
Though successful, these supervised methods highly rely on massive data samples with annotated action labels, which are expensive to obtain. 
Semi-supervised approaches are thus developed to alleviate this data annotation limitation, and in this paper, we apply it to learning motion representation for 3D action recognition. 

\subsection{Semi-Supervised Learning}

Semi-supervised learning algorithms learn from a data set that includes both labeled and unlabeled data, usually mostly unlabeled. For a comprehensive review of semi-supervised methods, we refer readers to \cite{Olivier2006Semi-Supervised}. Recently, there is increasing interest in deep learning based semi-supervised algorithms. One group of these methods is based on generative models, \eg denoising autoencoders \cite{Antti2005Semi-supervised}, variational autoencoders \cite{Durk2014Semi-supervised} and generative adversarial networks \cite{Augustus2016Semi-supervised,Tim2016Improved}. 
Some semi-supervised methods add small perturbations to unlabeled data, and require similar outputs between them by enforcing a consistency regularization, \eg $\rm\Pi$-Model  \cite{Samuli2017Temporal}, Temporal Ensembling \cite{Samuli2017Temporal}, Mean Teacher \cite{Antti2017Mean} and 
Virtual Adversarial Training \cite{Takeru2018Virtual}. 
There are also some other works. To name a few, Lee \etal \cite{Dong2013Pseudo} pick up the class with maximum predicted probability as pseudo-labels for unlabeled data, and use them to train the models. 
\cite{Yves2005Semi-supervised} presents a conditional entropy minimization for unlabeled data, which encourages their predicted probability to bias some class. 
The work most related to ours is \cite{Xiaohua2019S4l} which proposes a new technique for semi-supervised learning by leveraging SSL techniques to learn representation of unlabeled images. Their work enlarges the generalization of semi-supervised learning methods. 
In this work, we exploit effective SSL to learn discriminative motion representation for semi-supervised 3D action recognition. Moreover, we further propose a novel adversarial regularization to couple SSL into the semi-supervised algorithm.

\subsection{Self-Supervised Learning for Action Recognition}

Self-supervised learning for action recognition aims to learn motion representations from the unlabeled data by solving the pretext tasks. Recently,  a stream of studies \cite{Nitish2015Unsupervised,Ishan2016Shuffle,Basura2017Self-supervised,Hsin-Ying2017Unsupervised,Uta2018Improving,Dejing2019Self} design various temporal-related tasks to learn the temporal pattern from the unlabeled RGB videos. For example, a sequence sorting task is introduced in \cite{Hsin-Ying2017Unsupervised}. \cite{Zelun2017Unsupervised,Jiangliu2019Self-supervised} propose to learn the video representation by predicting motion flows. Note that, these methods are for learning representations from RGB videos and not applicable to long-term skeleton sequences. For 3D action recognition, Zheng \etal~\cite{Nenggan2018Unsupervised} propose a conditional skeleton inpainting architecture to learn the long-term dynamics from unlabeled skeleton data. However, this SSL ignores the shared information among samples with the same action class and therefore may yield less discriminative feature representations. Hence, we propose an effective self-supervised strategy to learn discriminative representation that is beneficial for semi-supervised 3D action recognition.

\section{Method}

\subsection{Problem Formulation}

Instead of relying on massive labels in existing methods,
we use only a few labeling data in semi-supervised 3D action recognition. 
Formally, let $\mathcal{X}$ be the training set. The training samples $\boldsymbol{x}_i \in \mathcal{X}$ are skeleton sequences with $T$ frames, and $\boldsymbol{x}_i = \{\boldsymbol{x}_{i,1},...,\boldsymbol{x}_{i,T} \}$. 
At each time $t$, the $\boldsymbol{x}_{i,t} $ is a set of 3D coordinates of body joints, which can be obtained by the Microsoft Kinect and the advanced human pose estimation algorithms \cite{cao2017realtime,Jianfeng2020Inference}. 
In contrast to supervised 3D action classification, training samples are split to two subsets in our task here:
a labeled training set denoted as $\mathcal{X}_L = \{ \boldsymbol{x}_1, ..., \boldsymbol{x}_L \}$   and
an unlabeled training set denoted as $\mathcal{X}_U = \{ \boldsymbol{x}_1, ..., \boldsymbol{x}_U \}$.
The training samples $\boldsymbol{x}_l \in \mathcal{X}_L$ have annotated labels $\{y_1, ..., y_L \}$ with $y_l \in  \mathcal{C}$, where $\mathcal{C} = \{1, ..., C \}$ is a discrete label set for $C$ action classes. The training samples $\boldsymbol{x}_u \in \mathcal{X}_U$  are unlabeled. Usually, $L$ is smaller than $U$ ($L \ll U$). 

Inspired by $S^4$L~\cite{Xiaohua2019S4l}, we propose an Adversarial Self-Supervised Learning framework to learn discriminative motion representations from $\mathcal{X}_L$ and $\mathcal{X}_U$. It couples self-supervised techniques into the semi-supervised scheme via adversarial learning and neighbor relation exploration. Detailed descriptions of ASSL are described in the following subsections.

\begin{figure}[t]
	\centering
	\includegraphics[width=0.75\linewidth, height=0.48\linewidth]{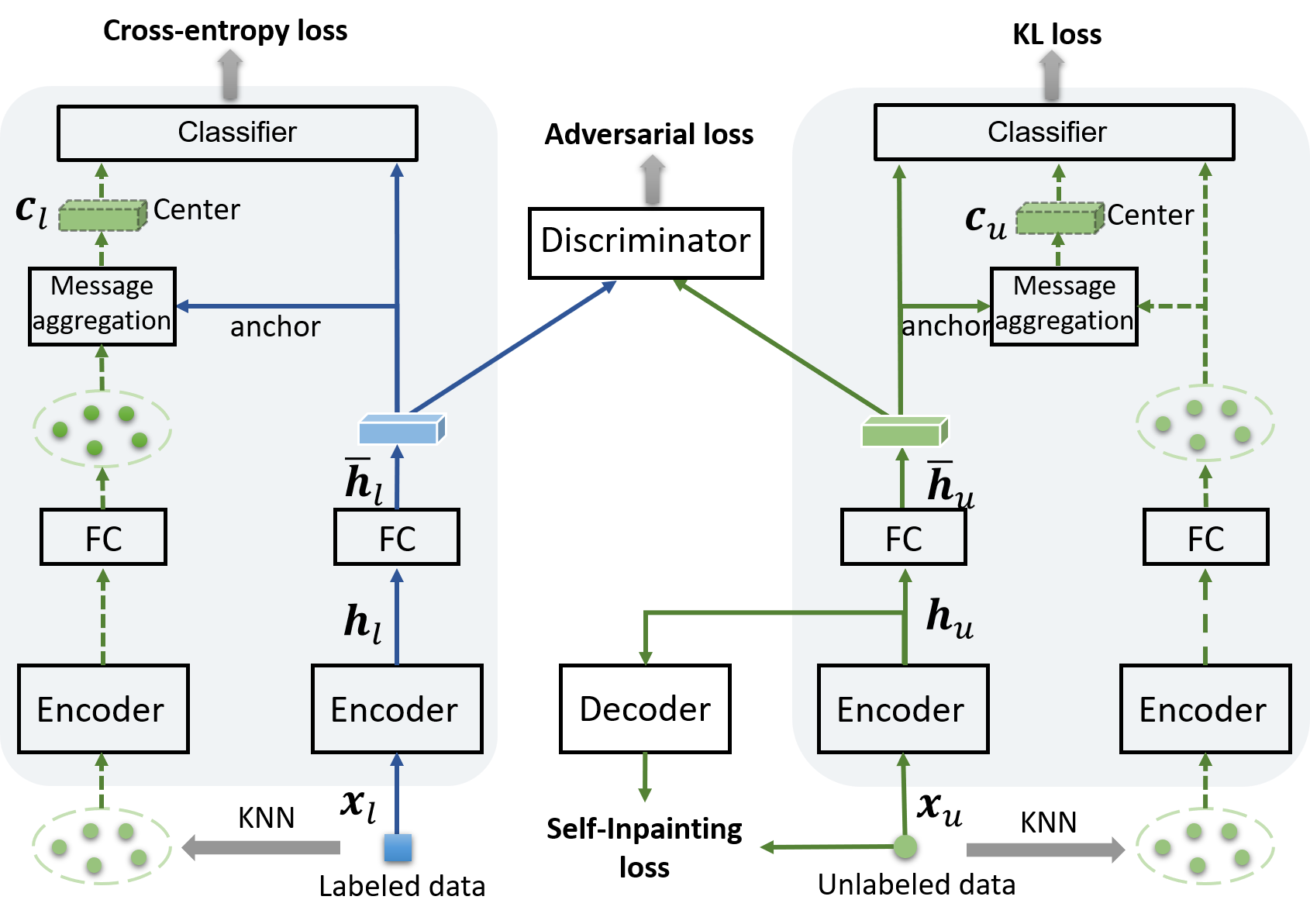}
   	\caption{Framework of Adversarial Self-Supervised Learning (ASSL).  The ASSL leverages SSL and adversarial regularization for semi-supervised 3D action recognition. For SSL techniques, in addition to a self-inpainting constraint \cite{Nenggan2018Unsupervised} for learning temporal information of each individual unlabeled sample, we propose to apply a new consistency regularization within the neighborhood to explore data relations. The adversarial training with a feature discriminator is used to align feature distributions of labeled and unlabeled samples, which further boosts generalization of semi-supervised models to unseen samples 
   }
   \label{fig_network}
\end{figure}

\subsection{Neighborhood Consistency for Semi-Supervised 3D Action Recognition}
\label{sec_selfsupervised}

Semi-supervised 3D action recognition aims to learn discriminative motion representation from massive unlabeled sequences. However, this is difficult over succinct 3D human poses. To tackle this challenge, we propose an effective SSL strategy, neighborhood consistency, that enhances the underlying class semantics of motion representation by exploring data relations within the neighborhoods, so as to improve recognition performance.

As shown in Fig. \ref{fig_network}, we first employ skeleton inpainting \cite{Nenggan2018Unsupervised} to learn temporal information for each unlabeled sequence.  Specifically, an encoder network $Enc$ takes an input skeleton sequence $\boldsymbol{x}_u$ from training set $\mathcal{X}_U$ and produces a vector as the temporal features $\boldsymbol{h}_u \in \mathbb{R}^d$. Conditioned on the learned representation $\boldsymbol{h}_u$, a decoder network $Dec$ aims to fill the masked regions in the input sequence. Due to the difference between the action classification (discrimination) and skeleton inpainting (regression) tasks, we use a translation layer \ie a linear layer, to bridge the gap between the feature spaces of both tasks. 
The output of linear layer is denoted as $\boldsymbol{\bar{h}}_u$ for the sample $\boldsymbol{x}_u$. Then, in this feature space, we employ K-nearest neighbor \cite{Thomas1967Nearest} to select $K$ nearest neighbors from unlabeled training set $\mathcal{X}_U$. The neighbor set of $\boldsymbol{x}_u$ is denoted as $\Omega_{x_u} = \{ \boldsymbol{x}^1_u, ...,\boldsymbol{x}^K_u \}$. A message aggregation module is proposed to produce the local center vector. We use a multilayer perceptron to assign a weight for each neighbor sample, which evaluates their similarities as the anchor. The weights $\alpha_k$ are computed as follows:
\begin{align}
    \label{eqn_weight}
  \alpha_k = \dfrac{ \exp \left( MLP \left( | \boldsymbol{\bar{h}}_u - \boldsymbol{\bar{h}}_u^k | \right) \right) }{ \sum_{k=1}^K \exp \left( MLP \left( | \boldsymbol{\bar{h}}_u - \boldsymbol{\bar{h}}_u^k |  \right) \right) },
\end{align}
where $ \boldsymbol{\bar{h}}_u^k$ is the translated feature of neighbor sample $\boldsymbol{x}^k_u \in \Omega_{x_u}$, $MLP(\cdot)$ denotes the multilayer perceptron in message aggregation module. According to the computed weights $\{\alpha_1, ..., \alpha_K\}$, the local class center $\boldsymbol{c}_u$ can be aggregated with the neighbor set $\Omega_{x_u}$ as follows:
\begin{align}
    \label{eqn_center}
    \boldsymbol{c}_u =  \sum_{k=1}^K \alpha_k \boldsymbol{\bar{h}}_u^k.
\end{align}

Considering the high compactness and class consistency within neighborhoods, we require that the samples within neighborhoods achieve a similar prediction with the local center $\boldsymbol{c}_u$. However, for a sample $\boldsymbol{x}_u$, its neighbor samples either share the class label (positive) with $\boldsymbol{x}_u$ or not (negative). To minimize the impact of negative neighbors, we introduce a simple selecting criterion: we get the 1-nearest labeled neighbor from the labeled training set $\mathcal{X}_L$ for the anchor $\boldsymbol{x}_u$ and the neighbor $\boldsymbol{x}_u^k$. If the labeled neighbors of the anchor $\boldsymbol{x}_u$ and the neighbor $\boldsymbol{x}_u^k$ have the same label, $\boldsymbol{x}_u^k$ is regarded as the positive neighbor. The set of selected positive neighbor for sample $\boldsymbol{x}_u$ is denoted as $\Omega_{x_u}^p$. Finally, the loss of consistency regularization within neighborhood is defined as follows: 
\begin{align}
\small
\label{eqn_KL1}
    \mathcal{L}_{KL} =  \sum_{\boldsymbol{x}_u \in \mathcal{X}_U} \left( KL\left( f_c(\boldsymbol{c}_u), f_c(\boldsymbol{\bar{h}}_u) \right) 
    + \sum_{x_u^K \in \Omega_{x_u}^p } KL \left( f_c(\boldsymbol{c}_u), f_c(\boldsymbol{\bar{h}}_u^k) \right) \right),
\end{align}
where $f_c(\cdot)$ is the classifier that outputs the predictions, $KL(\cdot)$ denotes Kullback-Leibler divergence. 

Like consistency regularization for unlabeled samples $\boldsymbol{x}_u \in \mathcal{X}_U$, the neighbor sets of labeled examples $\boldsymbol{x}_l \in \mathcal{X}_L$ are also selected from the unlabeled set $\mathcal{X}_U$. which are denoted as $\Omega_{\boldsymbol{x}_l}$. Similarly, we use the feature $\boldsymbol{\bar{h}}_l$ of $\boldsymbol{x}_l$ as the anchor to estimate its local center representation $\boldsymbol{c}_l$ with its neighbors set $\Omega_{\boldsymbol{x}_l}$ as the Eqn. \eqref{eqn_weight}-\eqref{eqn_center} (shown in Fig.~\ref{fig_network}). Under the assumption that the anchor shares the same class semantic with the local center, we use a cross-entropy loss $CE(\cdot)$  for the center $\boldsymbol{c}_l$:
\begin{align}
\label{eqn_KL2}
    \mathcal{L}_{{CE}}^{c} =  \sum_{\boldsymbol{x}_l \in \mathcal{X}_L} \left(  CE \left( f_c(\boldsymbol{c}_l), y_l \right) \right),
\end{align}
where $y_l$ is the class label of $\boldsymbol{x}_l$.

Overall, the optimization objectives of unlabeled samples can be formulated as follows:
\begin{align}
    \mathcal{L}_{U} =  \mathcal{L}_{KL} + \mathcal{L}_{{CE}}^{c} + \mathcal{L}_{inp},
    \label{eqn_unlabel}
\end{align}
where $\mathcal{L}_{inp}$ denotes the skeleton inpainting loss that is the $L_2$ distance between the inpainted sequence and the original input sequence. Minimizing this optimization objective $\mathcal{L}_{U}$ encourages the model to enhance the underlying class concept of the self-supervised motion representation and yield discriminative feature representations.

\subsection{Adversarial Learning for Aligning Self-Supervised and Semi-Supervised Representations}
\label{sec_Adversarial}

\begin{wrapfigure}{R}{0.5\textwidth}
  \centering
  \subfigure[\fontsize{7}{7}\selectfont Sup.]{
        \label{fig_feature_ntu:a}
        \includegraphics[width=0.9in]{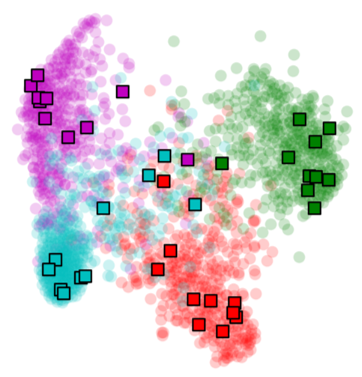}}
    \subfigure[\fontsize{7}{7}\selectfont Sup. + Sel.]{
        \label{fig_feature_ntu:b}
        \includegraphics[width=0.9in]{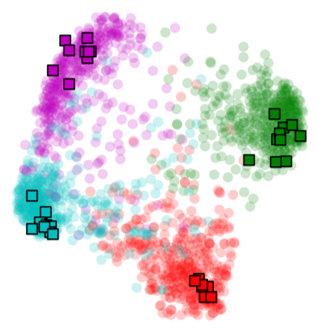}}
  \caption{The t-SNE visualization of motion features learned by \emph{Sup.} and \emph{Sup. + Sel.}. (a) \emph{Sup.} is trained with the supervised objective for the labeled samples. (b) \emph{Sup. + Sel.} is trained through optimizing the supervised and SSL objectives (Eqn.~\eqref{eqn_unlabel}) for the labeled and unlabeled samples, respectively. Different colors indicate different classes. Best viewed in color. The squares with black border denote the labeled data, and others are unlabeled data
  }
  \label{fig_feature_ntu}
\end{wrapfigure}
According to the training of existing semi-supervised learning methods, the labeled and unlabeled samples are enforced with  supervised and SSL objectives, respectively. In this work, Eqn.~\eqref{eqn_unlabel} is used for the unlabeled samples. Although our proposed SSL technique is quite effective for semi-supervised 3D action recognition, we identify that the representations learned with supervised and SSL task are misaligned. As shown in the Fig. \ref{fig_feature_ntu}, with the benefit of SSL technique, the features of \emph{Sup. + Sel.} present a more compact distribution than \emph{Sup.}. However, in contrast to the intra-class compactness of labeled data (the squares with black border), there are scattering distributions for the unlabeled data in Fig. \ref{fig_feature_ntu:b}. Thus, although both sequences are sampled from the same data distribution, their feature distributions are misaligned due to different optimization objectives. To tackle this problem, we propose a novel adversarial training strategy to couple SSL method with the semi-supervised 3D action recognition. Specifically, a discriminator $Dis$ is trained to distinguish the unlabeled features from the labeled features. And the model is trained simultaneously to confuse the discriminator $Dis$. Hence, the adversarial loss is defined as follows:
\begin{align}
    \mathcal{L}_{adv} =  \frac{1}{L} \sum_{\boldsymbol{x}_l \in \mathcal{X}_L} \left( \log \left( Dis (\boldsymbol{\bar{h}}_l) \right)  \right) + \frac{1}{U} \sum_{\boldsymbol{x}_u \in \mathcal{X}_U} \left( \log \left( 1 - Dis(\boldsymbol{\bar{h}}_u) \right) \right).
\end{align}

The adversarial regularization allows the model to align the feature distributions of labeled and unlabeled data. Therefore, like the labeled data, the feature distribution of unlabeled data becomes more intra-class compactness, which boosts the capability of generalization to unseen samples. More analyses about adversarial regularization are reported in Section \ref{sec_adversarial_analysis}.

\subsection{Model Architecture and Optimization}
\label{sec_Training}

Unlike the existing 3D action recognition method \cite{Du2015Hierarchical,Zhang2017View,Chao2018Co-occurrence,Ke2017A,Yan2018Spatial,Shi_2019_CVPR,Si_2019_CVPR} learning the discriminative features through the designed effective networks, the goal of this work is to explore effective semi-supervised scheme for 3D action recognition. Therefore, this work adopts a universal architecture. In order to effectively capture the motion dynamics, we use three bidirectional GRU layers to encode the input skeleton sequence in the $Enc$. The decoder consists of two unidirectional GRU layers. We use 4 linear layers and 3 linear layers in the discriminator and the multilayer perceptron of message aggregation, respectively. The classifier is a two-layer perceptron. 

During training, our ASSL network is learned by minimizing the following loss on the training data:
\begin{align}
    \label{eqn_loss}
    \mathcal{L} = \mathcal{L}_{L} + \lambda_1 \mathcal{L}_{U} +  \lambda_2 \mathcal{L}_{adv}.
\end{align}
where $\mathcal{L}_L$ is a cross-entropy loss of all labeled examples in $\mathcal{X}_L$,  $\lambda_1$ and $\lambda_2$ are non-negative scalar weights. Note that, 
we always sample the same number labeled and unlabeled samples in mini-bathes.

\section{Experiments}

In this section, we evaluate and compare our work with previous semi-supervised methods and also conduct detailed component analysis.  

\subsection{Experimental Setup}

\paragraph{\textbf{Datasets}}
Two popular benchmark datasets, NTU RGB+D dataset \cite{Shahroudy2016NTU} and Northwestern-UCLA dataset \cite{Jiang2014Cross}, are used for our experiments. 

NTU RGB+D dataset \cite{Shahroudy2016NTU} contains 56,880 samples covering 60 different classes of human actions performed by 40 distinct subjects.
These videos are collected with three cameras simultaneously in different horizontal views. 
Two evaluation protocols are provided: Cross-Subject (CS) and Cross-View (CV). For CS protocol, skeleton sequences performed by 20 subjects are used for training, and the rest for testing. For CV protocol, all videos from Camera 2 and 3 are used for training while those from Camera 1 are used for testing. For semi-supervised 3D action recognition, 5\%, 10\%, 20\% and 40\% of training sequences of each class are labeled on the training set.

Northwestern-UCLA dataset \cite{Jiang2014Cross} has 1,494 samples performed by 10 different subjects belonging to 10 action classes. Each action sample is captured by three Kinect cameras simultaneously from a variety of viewpoints. 
Its training set consists of samples from the first two cameras and the rest from the third camera form the testing set. 
For semi-supervised 3D action recognition, 
we use 5\%, 15\%, 30\% and 40\% labels of training sequences of each class on the training set.

\paragraph{\textbf{Baselines}}
There is no available semi-supervised baseline for 3D action recognition, so we use following methods as baselines that achieve state-of-the-art performances in the RGB domain:

\begin{itemize}
  \item [1)] 
  Supervised-only (Sup.), training with labeled skeleton sequences only.    
  \item [2)]
  Pseudo labels \cite{Dong2013Pseudo}, leveraging the idea that the predicted labels of unlabeled samples are used for training. First, train a model with the labeled data, then predict the classes of unlabeled samples. These pseudo labels are used to retrain the network in a supervised fashion with labeled and unlabeled data simultaneously.
  \item [3)]
  Virtual Adversarial Training (VAT) \cite{Takeru2018Virtual}, training with unlabeled data to make the model robust around input data point against local perturbation. It generates small adversarial perturbations for unlabeled samples, which greatly alter the output distribution; then consistency loss is applied over unlabeled training data to encourage consistency of predictions for input data and its adversarially perturbed version.
    \item [4)]
  Conditional Entropy Minimization (EntMin) \cite{Yves2005Semi-supervised}, minimizing the entropy of prediction over unlabeled training data as a regularization for model training. Predicted class probabilities are encouraged to be near a one-hot vector via training with unlabeled data.
  \item [5)]
  Self-Supervised Semi-Supervised Learning ($S^4$L) \cite{Xiaohua2019S4l}, the most related method to ours. It trains the model on self-supervised and semi-supervised tasks in a multi-task fashion. For 3D action recognition, we use the skeleton inpainting framework \cite{Nenggan2018Unsupervised} as the pretext task for self-supervised learning.
\end{itemize}

\paragraph{\textbf{Implementation}}
All comparisons with semi-supervised baselines are made under the same setting to be fair. In all experiments, the dimension of hidden states in the GRU and bidirectional GRU is set to 512. On both datasets, we randomly sample $T=40$ frames from each skeleton sequence as input during training and testing. We train all networks by the ADAM optimizer \cite{kingma2015adam}. The learning rate, initiated with 0.0005, is reduced by multiplying it by 0.5 every 30 epochs. We set $\lambda_1 = 1$ and $\lambda_2 = 0.1$ in Eqn. \eqref{eqn_loss}.  Our experiments are all implemented with PyTorch and 1 Titan Xp GPU.

\subsection{Comparison with Semi-Supervised Methods}
\label{Comparison_result}

\begin{table}[b]
\begin{center}
\caption{ Test accuracy (\%) on NTU dataset (Cross-Subject (CS) and Cross-View protocols (CV)) with 5\%, 10\%, 20 and 40\% labels of training set.
\emph{v./c.} denotes the number of labeled videos per class 
}
\label{ntu_result}
\resizebox{1.0\linewidth}{!}{
\begin{tabular}{l|cc|cc|cc|cc}
\toprule \noalign{\smallskip}
    \multirow{2}{*}{Method} & \multicolumn{2}{c}{5\%} & \multicolumn{2}{|c}{10\% } & \multicolumn{2}{|c}{20\% } & \multicolumn{2}{|c}{40\% } \\
\noalign{\smallskip}
    \cline{2-9}
\noalign{\smallskip}
    ~ & \multicolumn{1}{c|}{CS (33 \emph{v./c.} )}  & \multicolumn{1}{c|}{CV (31 \emph{v./c.})} & \multicolumn{1}{c|}{CS (66 \emph{v./c.})} & \multicolumn{1}{c|}{CV (62 \emph{v./c.})} & \multicolumn{1}{c|}{CS (132 \emph{v./c.})} & \multicolumn{1}{c|}{CV (124 \emph{v./c.})} & \multicolumn{1}{c|}{CS (264 \emph{v./c.})} & \multicolumn{1}{c}{CV (248 \emph{v./c.})} \\
\midrule
    Supervised-only    & 47.2 & 53.7 & 57.2 & 63.1 & 62.4 & 70.4 & 68.0 & 76.8 \\
\midrule
    Pseudolabels \cite{Dong2013Pseudo}              & 50.9 & 56.3 & 58.4 & 65.8 & 63.9 & 71.2 & 69.5 & 77.7 \\
    VAT \cite{Takeru2018Virtual}                    & 51.3 & 57.9 & 60.3 & 66.3 & 65.6 & 72.6 & 70.4 & 78.6 \\
    VAT + EntMin \cite{Yves2005Semi-supervised}     & 51.7 & 58.3 & 61.4 & 67.5 & 65.9 & 73.3 & 70.8 & 78.9 \\
    $S^4$L (Inpainting) \cite{Xiaohua2019S4l}       & 48.4 & 55.1 & 58.1 & 63.6 & 63.1 & 71.1 & 68.2 & 76.9 \\
\midrule
    ASSL (ours)                             & \textbf{57.3} & \textbf{63.6} & \textbf{64.3} & \textbf{69.8} & \textbf{68.0} & \textbf{74.7} & \textbf{72.3} & \textbf{80.0} \\
\bottomrule
\end{tabular}
}
\end{center}
\end{table}

We evaluate our method by comparing it with baselines for semi-supervised 3D action recognition and show results on NTU and N-UCLA datasets respectively in
Tables \ref{ntu_result} and \ref{NUCLA_result}.

As seen from tables, with the proposed ASSL network, we establish new state-of-the-art performances of semi-supervised 3D action recognition. To be specific, \emph{$S^4$L (Inpainting)} performs worse than \emph{Pseudolabels}, \emph{VAT} and \emph{VAT + EntMin}, suggesting it is inefficient to learn discriminative representation via skeleton inpainting and thus semi-supervised 3D action recognition has derived little benefit from self-supervised representations. \emph{$S^4$L (Inpainting)}, though a advanced semi-supervised approach, requires an effective self-supervised representations that are difficult to learn in this task. Compared with these semi-supervised methods, our benefit is larger when the number of labels is reduced. For example, with 5\% labels of training set on NTU dataset, the results of our ASSL present greater improvement compared with \emph{VAT + EntMin}. This clearly demonstrates the power of the proposed ASSL.

\begin{table}[t]
\begin{center}
\caption{Test accuracy (\%) on N-UCLA dataset with 5\%, 15\%, 30\% and 40\% labels of training set. \emph{v./c.} denotes the number of labeled videos per class}
\label{NUCLA_result}
\resizebox{0.7\linewidth}{!}{
\begin{tabular}{l|cccc}
\toprule \noalign{\smallskip}
    Method & 5\% (5 \emph{v./c.}) & 15\% (15 \emph{v./c.})   & 30\% (30 \emph{v./c.}) & 40\% (40 \emph{v./c.})  \\

\midrule
    Supervised-only                                 & 34.1 & 37.9  & 48.9 & 58.8  \\
\midrule
    Pseudolabels \cite{Dong2013Pseudo}              & 35.6 & 48.9  & 60.6 & 65.7  \\
    VAT \cite{Takeru2018Virtual}                    & 44.8 & 63.8  & 73.7 & 73.9  \\
    VAT + EntMin \cite{Yves2005Semi-supervised}     & 46.8 & 66.2  & 75.4 & 75.6   \\
    $S^4$L (Inpainting) \cite{Xiaohua2019S4l}       & 35.3 & 46.6  & 54.5 & 60.6   \\
\midrule
    ASSL (ours)                             & \textbf{52.6} & \textbf{74.8} & \textbf{78.0} & \textbf{78.4} \\
\bottomrule
\end{tabular}
}
\end{center}
\end{table}

\subsection{Ablation Study}
\label{sec_ablation}

We then investigate effectiveness of the neighborhood consistency and adversarial training in our proposed ASSL on NTU and N-UCLA datasets. We also analyze effects of different neighborhood sizes and Neighborhood quality.

\paragraph{\textbf{Neighborhood Consistency}}
We evaluate the effects of the proposed self-supervised strategy, neighborhood consistency, upon the discriminativeness of motion representations that is shown in final performance of semi-supervised 3D action recognition. 
In Table \ref{analysis_result}, the model \emph{Sup. + Inp.} is trained with a cross-entropy loss for labeled data and a self-inpainting loss $\mathcal{L}_{inp}$  for unlabeled data. 
Instead of self-inpainting loss, \emph{Sup. + Nei.} explores the data relations within neighborhoods by enforcing the consistency regularization (Eqn. \eqref{eqn_KL1}, \eqref{eqn_KL2}) for unlabeled data. 
We can see that \emph{Sup. + Nei.} significantly outperforms the \emph{Sup. + Inp.}. The comparison  results justify that our neighborhood consistency can learn more discriminative motion representations that are more beneficial for semi-supervised 3D action recognition. 

Moreover, the self-inpainting constraint \cite{Nenggan2018Unsupervised} aims at learning temporal information of each \emph{individual} unlabeled sequence. 
The goal of our neighborhood consistency regularization is to explore inter-sample relations within neighborhoods. 
We jointly learn the two features in \emph{Sup. + Inp. + Nei.}. 
It can be seen compared with \emph{Sup. + Inp.} and \emph{Sup. + Nei.}, \emph{Sup. + Inp. + Nei.} achieves better performance on both datasets for semi-supervised 3D action recognition. This illustrates that the representations learned by our neighborhood consistency are complementary to those learned with self-inpainting. Therefore, the benefits of combining these two SSL techniques to capture discriminative representation from unlabeled sequences in our final model are verified (seen Eqn. \eqref{eqn_unlabel}).

\begin{table}[t]
\begin{center}
\caption{Ablation study on self-supervised learning methods,  skeleton inpainting (\emph{Inp.}) \cite{Nenggan2018Unsupervised} and neighbor consistency (\emph{Nei.}). 
Classification accuracy (\%) is reported on NTU with 5\% labels and N-UCLA with 15\% labels. 
}
\label{analysis_result}
\resizebox{0.6\linewidth}{!}{
\begin{tabular}{l|cc|c}
\toprule \noalign{\smallskip}
    \multirow{2}{*}{Methods} & \multicolumn{2}{c|}{NTU 5\% }  & N-UCLA  15\%   \\
\cline{2-3}
    ~                        & \multicolumn{1}{c|}{CS (33 \emph{v./c.})}  & \multicolumn{1}{c|}{CV (31 \emph{v./c.})}   & (15 \emph{v./c.}) \\
\midrule
    Supervised-only (Sup.)                & 47.2 & 53.7 & 37.9 \\
\midrule
    Sup. + Inp.                                & 48.4 & 55.1 & 46.6  \\
    Sup. + Nei.                                & 52.1 & 57.8 & 60.0  \\
    Sup. + Inp. + Nei.                         & 55.2 & 61.1 &  66.4  \\
\midrule
    ASSL                             & \textbf{57.3} & \textbf{63.6} &  \textbf{74.8}   \\
\bottomrule
\end{tabular}
}
\end{center}
\end{table}

\begin{figure}[b]
    \begin{minipage}[t]{0.49\linewidth} 
    \centering 
        \includegraphics[width=2.3in, height=1.2in]{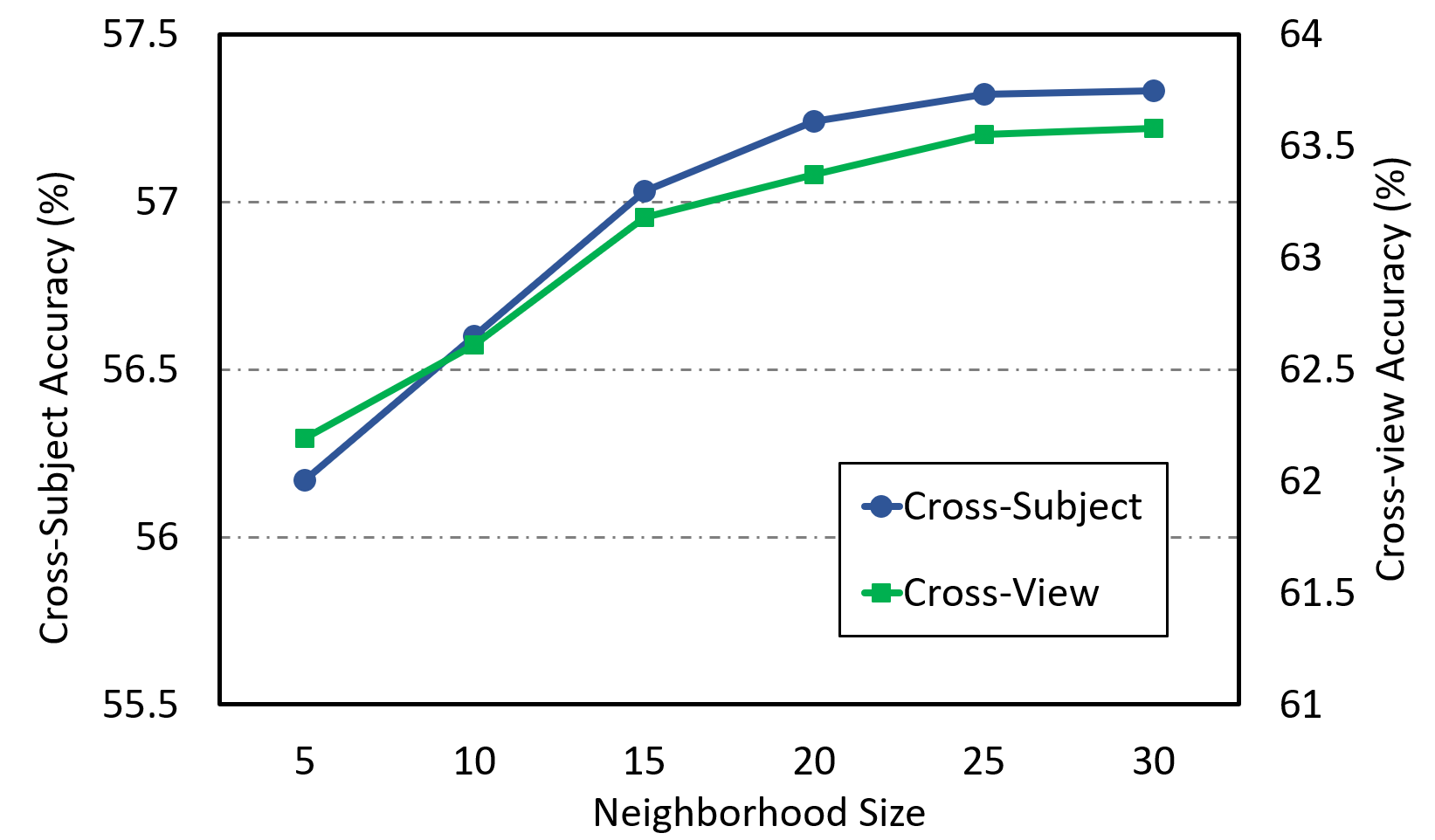} 
        \caption{Classification accuracy (\%) with different neighborhood size on NTU dataset with 5\% labels 
        } 
    \label{fig_neighbor} 
    \end{minipage}%
    \hspace{1mm}
    \begin{minipage}[t]{0.49\linewidth} 
    \centering 
        \includegraphics[width=2.1in, height=1.2in]{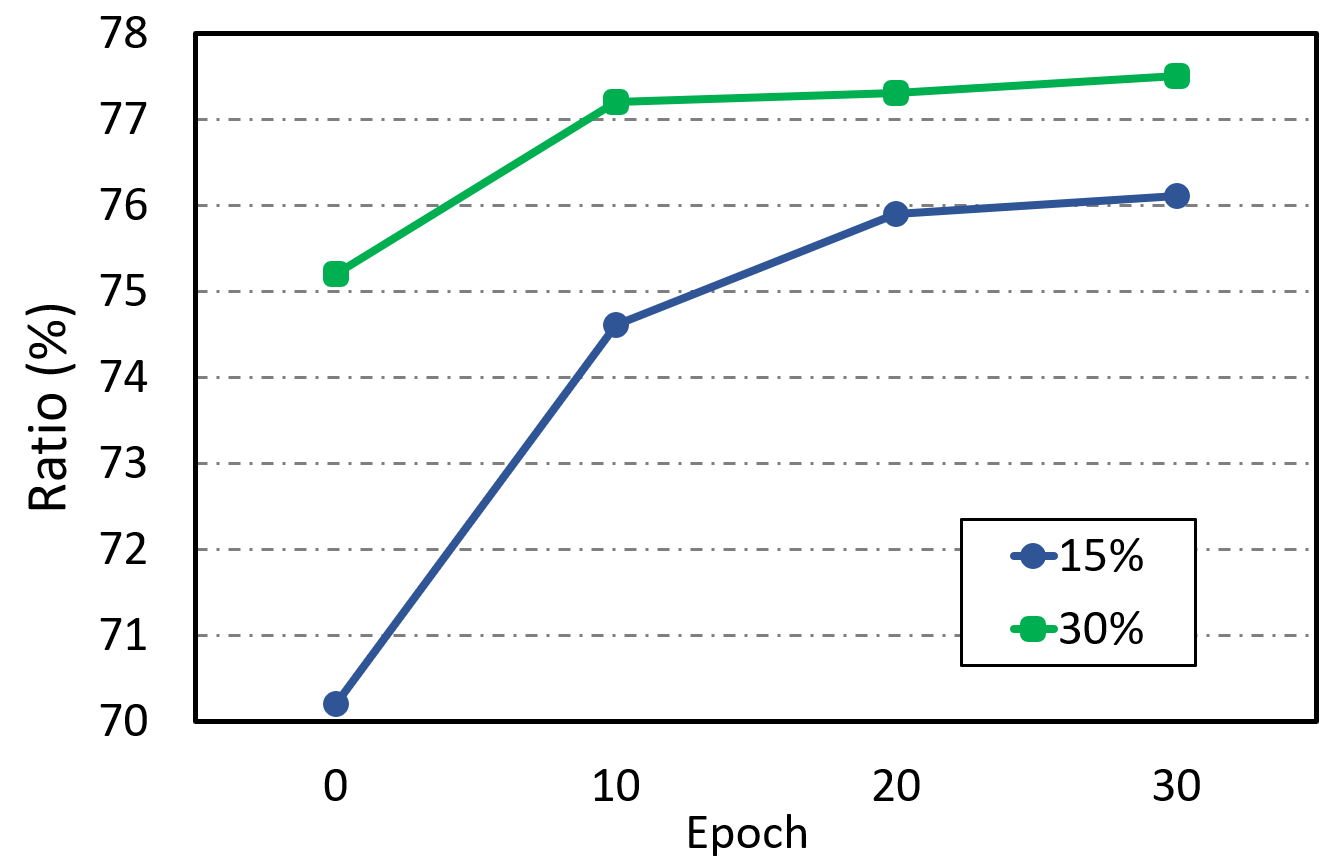} 
        \caption{The ratio of neighbor samples sharing the same action label as the anchor throughout training on N-UCLA dataset} 
    \label{fig_neighboquality} 
    \end{minipage} 
\end{figure}

\paragraph{\textbf{Neighborhood Size}}
We assume that the larger neighborhood size imposes stronger regularization and gives better performance. In order to justify this hypothesis, we investigate the effects of different neighborhood sizes in Fig. \ref{fig_neighbor}. As neighborhood size increases, the performance is improved and then becomes saturated. 
This implies that more discriminative representations can be learned with a larger size. 
But, if using too large a size, the model will cover distant data points that have weak semantic consistency within the neighborhood,and hence the performance becomes saturated.

\paragraph{\textbf{Neighborhood Quality}}
We further examine effects of the class consistency of anchor Neighborhood, \ie Neighborhood quality. 
In Fig. \ref{fig_neighboquality}, we report the progress of the ratio of neighbor samples sharing the same action label as the anchor throughout training. 
We can observe the ratio of class consistent neighborhoods increases, and then becomes saturated. 
This indicates exploring data relations is helpful to inferring  underlying class semantics, thus facilitating the clustering of samples with the same action labels.

\begin{table}[t]
\begin{center}
\caption{Ablation study on adversarial training. Classification accuracy (\%) is reported on NTU with 5\% labels and N-UCLA with 15\% labels. 
}
\label{analysis_adv}
\resizebox{0.65\linewidth}{!}{
\begin{tabular}{cc|cc|c}
\toprule \noalign{\smallskip}
    \multicolumn{2}{c|}{\multirow{2}{*}{Methods}} & \multicolumn{2}{c|}{NTU 5\% }  & N-UCLA  15\%   \\
    \cline{3-4}
    \multicolumn{2}{c|}{}                        & \multicolumn{1}{c|}{CS (33 \emph{v./c.})} & \multicolumn{1}{c|}{CV (31 \emph{v./c.})}  & (15 \emph{v./c.}) \\
\midrule

    \multirow{2}{*}{Sup. + Inp.}               & \emph{w/o} adv & 48.4 & 55.1 & 46.6  \\
    ~                                                & \emph{w/} adv & 51.2 & 57.1 & 52.4  \\
    \midrule
    \multirow{2}{*}{Sup. + Nei.}                    & \emph{w/o} adv & 52.1 & 57.8 & 60.0  \\
    ~                                                & \emph{w/} adv & 53.4 & 59.1 & 68.5  \\
    \midrule
    ASSL                                           & \emph{w/o} adv & 55.2 & 61.1 & 66.4  \\
    (Sup. + Inp. + Nei.)                                                 & \emph{w/} adv & 57.3 & 63.6 & 74.8  \\

\bottomrule
\end{tabular}
}
\end{center}
\end{table}

\begin{figure}[!b]
	\centering
	\subfigure[\fontsize{6}{6}\selectfont CS-Sup.]{
        \label{fig_feature:a}
        \includegraphics[width=1.01in, height=0.91in]{./pic/cs_baseline.png}}
    \hspace{10mm}
    \subfigure[\fontsize{6}{6}\selectfont CS-ASSL \emph{w/o} adv ]{
        \label{fig_feature:b}
        \includegraphics[width=1.0in, height=0.91in]{./pic/cs_woadv.png}}
    \hspace{10mm}
    \subfigure[\fontsize{6}{6}\selectfont CS-ASSL\emph{w/} adv]{
        \label{fig_feature:c}
        \includegraphics[width=1.0in, height=0.91in]{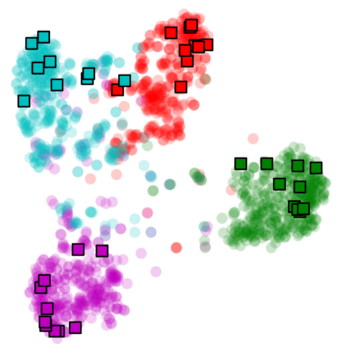}}

    \subfigure[\fontsize{6}{6}\selectfont CV-Sup.]{
        \label{fig_feature:d}
        \includegraphics[width=1.0in, height=0.91in]{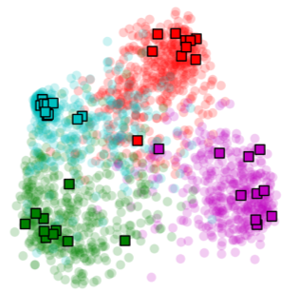}}
    \hspace{10mm}
    \subfigure[\fontsize{6}{6}\selectfont CV-ASSL \emph{w/o} adv]{
        \label{fig_feature:e}
        \includegraphics[width=1.0in, height=0.91in]{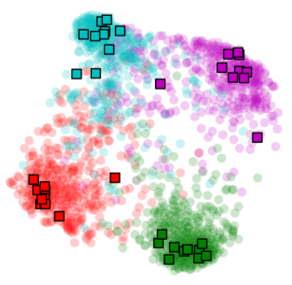}}
    \hspace{10mm}
    \subfigure[\fontsize{6}{6}\selectfont CV-ASSL \emph{w/} adv]{
        \label{fig_feature:f}
        \includegraphics[width=1.0in, height=0.91in]{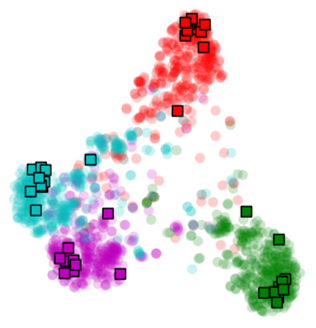}}
   	\caption{The t-SNE visualization of motion features learned by \emph{Supervised Baseline} (Sup.), \emph{ASSL w/o adv} and \emph{ASSL w/ adv} (ours) on NTU dataset. Different colors indicate different classes. Best viewed in color. The squares with black border denote the labeled data, and others are unlabeled data  }
   	\label{fig_feature}
\end{figure}

\paragraph{\textbf{Adversarial Training}}
\label{sec_adversarial_analysis}
The adversarial alignment is proposed to mitigate the gap between representations learned from supervised and self-supervised tasks. 
To evaluate effectiveness of adversarial training for coupling self-supervision methods with the semi-supervised 3D action recognition, we train several self-supervised models with or without adversarial regularization. 
The results are reported in Table \ref{analysis_adv}. 
It is obvious that all models with adversarial regularization achieve better performances than those without. 
For example, on N-UCLA dataset, the result of \emph{ASSL w/ adv} is 74.8\%, outperforming   \emph{ASSL w/o adv} by 8.4\%. 
The improved performance in Table \ref{analysis_adv} demonstrates that it is an effective strategy to couple self-supervision  with semi-supervised algorithms by adversarial training.

To further explore this scheme, we visualize the feature distributions of \emph{Sup.}, \emph{ASSL w/o adv} and \emph{ASSL w/ adv} by using t-SNE \cite{Visualizing2008Visualizing} in Fig. \ref{fig_feature}. 
For the model \emph{Sup.} trained with only supervised objective on labeled data, the decision boundaries of its feature distributions are very ambiguous.  The model \emph{ASSL w/o adv} is trained with supervised and self-supervised objectives for labeled and unlabeled data, respectively. 
Compared with \emph{Sup.}, the features of \emph{ASSL w/o adv} present tighter distributions, which benefit from self-supervised learning. 
But, long-tail distributions still exist for unlabeled samples (circles). 
Fig. \ref{fig_feature:c} and \ref{fig_feature:f} show clearly the alignment between feature distributions of labeled and unlabeled data for \emph{ASSL w/ adv}, \ie the proposed ASSL.  Overall, the comparison results prove the effectiveness of adversarial training for coupling self-supervision with semi-supervised action recognition. And this drives our model to learn more discriminative features that have desired intra-class compactness and inter-class separability.

\section{Conclusions}
In this paper, we consider the semi-supervised learning scheme for 3D action recognition task. The proposed ASSL effectively couples SSL into semi-supervised algorithm via neighbor relation exploration and adversarial learning. Exploring data relations with neighborhood consistency regularization encourages the model to learn discriminative motion representations that significantly improve the performance of this task. Moreover, we introduce a novel adversarial regularization to couple SSL method into a semi-supervised algorithm. This allows the model to align the feature distributions of labeled and unlabeled samples and boosts the capability of generalization to unseen samples. Our experiments verify that the proposed neighbor relation exploration and adversarial learning are strongly beneficial for semi-supervised 3D action recognition. With the proposed ASSL network, we establish news state-of-the-art performances of semi-supervised 3D action recognition. 

\section*{Acknowledgements}
\label{ Acknowledgements}

This work is jointly supported by National Key Research and Development Program of China (2016YFB1001000), National Natural Science Foundation of China (61420106015, 61976214, 61721004), Shandong Provincial Key Research and Development Program (Major Scientific and Technological Innovation Project) (NO.2019JZZY010119). Jiashi Feng was partially supported by MOE Tier 2 MOE2017-T2-2-151, NUS\_ECRA\_FY17\_P08, AISG-100E-2019-035. Chenyang Si was partially supported by the program of China Scholarships Council (No.201904910608). We thank Jianfeng Zhang for his helpful comments.

\clearpage
%
%
\bibliographystyle{splncs04}
\bibliography{egbib}
\end{document}